\documentclass[11pt]{article}
\usepackage{paclic33}
\usepackage[authoryear,round]{natbib}
\usepackage[hidelinks]{hyperref}
\usepackage[utf8]{inputenc}
\usepackage{graphicx}
\usepackage{amsmath}
\usepackage{multirow}
\usepackage{booktabs}
\usepackage{algorithm}
\usepackage{algorithmic}
\usepackage{times}  %Required
\usepackage{helvet}  %Required\urlstyle{same}
\usepackage{courier}  %Required
\usepackage{url}  %Required% the following package is optional:
\usepackage{graphicx}  %Required%\usepackage{latexsym} 
\usepackage{subfigure}
\usepackage{bm}% Following comment is from ijcai97-submit.tex:
\usepackage{amsmath}% The preparation of these files was supported by Schlumberger Palo Alto
\usepackage{color}% Research, AT\&T Bell Laboratories, and Morgan Kaufmann Publishers.
\usepackage{booktabs}  % Shirley Jowell, of Morgan Kaufmann Publishers, and Peter F.
\usepackage{multirow}% Patel-Schneider, of AT\&T Bell Laboratories collaborated on their
\usepackage{setspace}% preparation.
\usepackage{CJK}
\usepackage{float}
\usepackage{amsfonts}% These instructions can be modified and used in other

\title{Explicit Contextual Semantics for Text Comprehension}

\author{
	Zhuosheng Zhang$^{1,2,3,}$\thanks{$\ $ These authors contribute equally. $\dagger$ Corresponding author. This paper was partially supported by National Key Research and Development Program of China (No. 2017YFB0304100) and Key Projects of National Natural Science Foundation of China (U1836222 and 61733011).}  ,
	Yuwei Wu$^{1,2,3,4,*}$,
	Zuchao Li$^{1,2,3}$,
	Hai Zhao$^{1,2,3,\dagger}$
	\\
	$^1$Department of Computer Science and Engineering, Shanghai Jiao Tong University\\
	$^2$Key Laboratory of Shanghai Education Commission for Intelligent Interaction\\
	and Cognitive Engineering, Shanghai Jiao Tong University, Shanghai, China\\
	$^3$MoE Key Lab of Artificial Intelligence, AI Institute, Shanghai Jiao Tong University, Shanghai, China\\
	$^{4}$College of Zhiyuan, Shanghai Jiao Tong University, China\\
	{\tt\{zhangzs,will8821,charlee\}@sjtu.edu.cn},
	{\tt zhaohai@cs.sjtu.edu.cn}
}

\date{}

\begin{document}
\maketitle
\begin{abstract}
  $\emph{Who did what to whom}$ is a major focus in natural language understanding, which is right the aim of semantic role labeling (SRL) task. Despite of sharing a lot of processing characteristics and even task purpose, it is surprisingly that jointly considering these two related tasks was never formally reported in previous work. Thus this paper makes the first attempt to let SRL enhance text comprehension and inference through specifying verbal predicates and their corresponding semantic roles. In terms of deep learning models, our embeddings are enhanced by explicit contextual semantic role labels for more fine-grained semantics. We show that the salient labels can be conveniently added to existing models and significantly improve deep learning models in challenging text comprehension tasks. Extensive experiments on benchmark machine reading comprehension and inference datasets verify that the proposed semantic learning helps our system reach new state-of-the-art over strong baselines which have been enhanced by well pretrained language models from the latest progress.
\end{abstract}

\section{Introduction}

Text comprehension is challenging for it requires computers to read and understand natural language texts to answer questions or make inference, which is indispensable for advanced context-oriented dialogue \citep{zhang2018DUA,Zhu2018lingke} and interactive systems \citep{chen2015neural,Huang2018Moon,zhang2019ime}. This paper focuses on two core text comprehension (TC) tasks, \emph{machine reading comprehension} (MRC) and \emph{textual entailment} (TE). 

One of the intrinsic challenges for text comprehension is semantic learning. Though deep learning has been applied to natural language processing (NLP) tasks with remarkable performance \citep{Cai2017Fast,zhang2018exploring,zhang2018OneShot,Bai2018deep,Zhang2018Effective,xiao-etal-2019-lattice}, recent studies have found deep learning models might not really understand the natural language texts \citep{Mudrakarta2018Did} and vulnerably suffer from adversarial attacks \citep{Jia2017Adversarial}. Typically, an MRC model pays great attention to non-significant words and ignores important ones. To help model better understand natural language, we are motivated to discover an effective way to distill semantics inside the input sentence explicitly, such as semantic role labeling, instead of completely relying on uncontrollable model parameter learning or manual pruning.

Semantic role labeling (SRL) is a shallow semantic parsing task aiming to discover \emph{who} did \emph{what} to \emph{whom}, \emph{when} and \emph{why}  \citep{He2018Syntax,li2018unified,li2019dependency}, providing explicit contextual semantics, which naturally matches the task target of text comprehension. For MRC, questions are usually formed with \emph{who}, \emph{what}, \emph{how}, \emph{when} and \emph{why}, whose predicate-argument relationship that is supposed to be from SRL is of the same importance as well. Besides, explicit semantics has been proved to be beneficial to a wide range of NLP tasks, including discourse relation sense classification \citep{Mihaylov2016Discourse}, machine translation \citep{Shi2016Knowledge} and question answering \citep{Yih2016The}. All the previous successful work indicates that explicit contextual semantics may hopefully help into reading comprehension and inference tasks.

Some work studied question answering (QA) driven SRL, like QA-SRL parsing \citep{he2015question,Mccann2018The,Fitzgerald2018Large}. They focus on detecting argument spans for a predicate and generating questions to annotate the semantic relationship. However, our task is quite different. In QA-SRL, the focus is commonly simple and short factoid questions that are less related to the context, let alone making inference. Actually, text comprehension and inference are quite challenging tasks in NLP, requiring to dig the deep semantics between the document and comprehensive question which are usually raised or re-written by humans, instead of shallow argument alignment around the same predicate in QA-SRL. In this work, to alleviate such an obvious shortcoming about semantics, we make attempt to explore integrative models for finer-grained text comprehension and inference.
\begin{figure}
	\centering
	\includegraphics[width=0.48\textwidth]{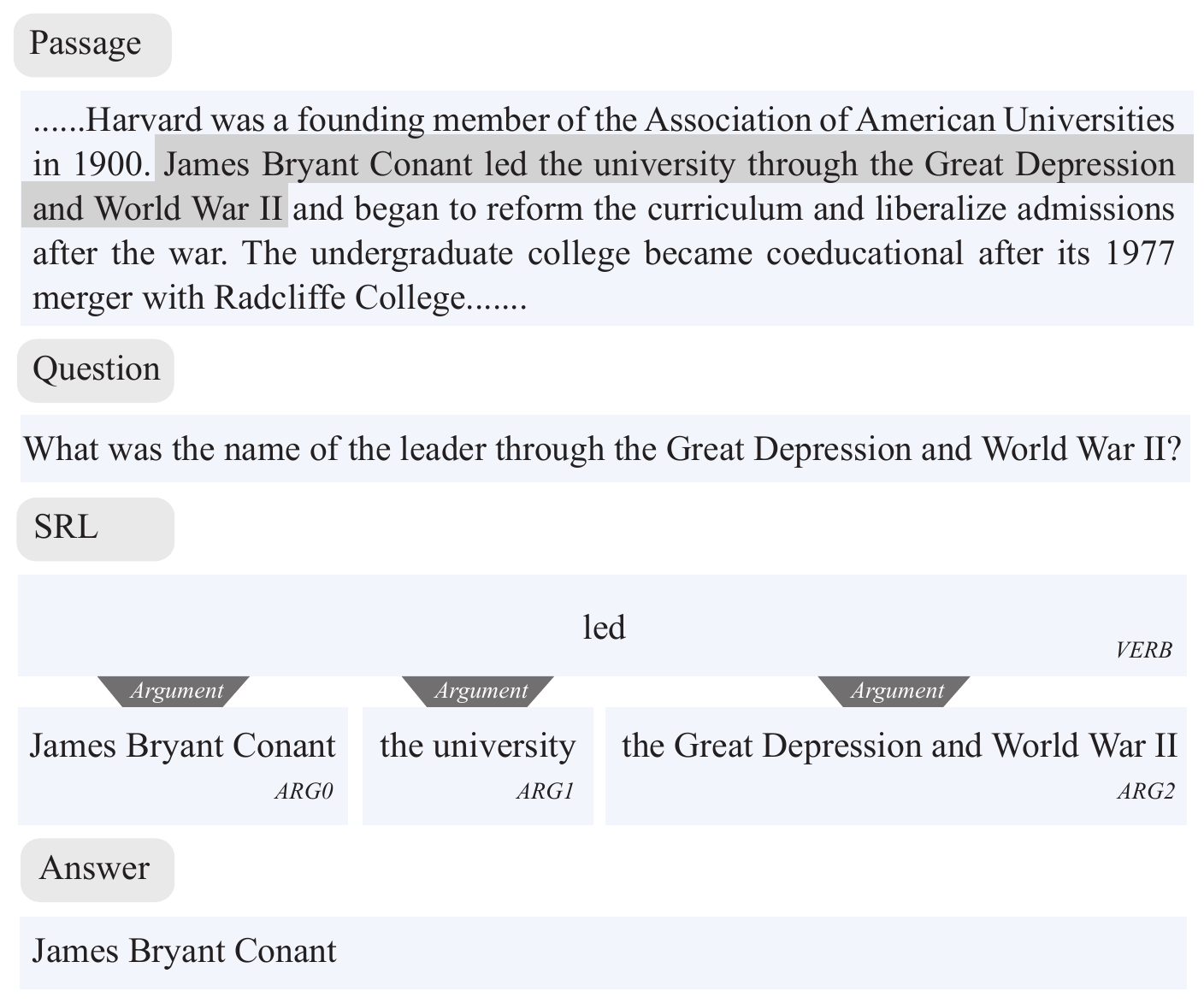}
	\caption{\label{fig:srl_example}Semantic role labeling guides text comprehension.}
\end{figure}

In this work, we propose a semantics enhancement framework for TC tasks, which boosts the strong baselines effectively. We implement an easy and feasible scheme to integrate semantic signals in downstream neural models in end-to-end manner to boost strong baselines effectively. An example about how contextual semantics helps MRC is illustrated in Figure \ref{fig:srl_example}. A series of detailed case studies are employed to analyze the robustness of the semantic role labeler. To our best knowledge, our work is the first attempt to apply explicit contextual semantics for text comprehension tasks, which have been ignored in previous works for a long time.

The rest of this paper is organized as follows. The next section reviews the related work. Section 3 will demonstrate our semantic learning framework and implementation. Task details and experimental results are reported in Section 4, followed by case studies and analysis in Section 5 and conclusion in Section 6.

\section{Related Work}

\subsection{Text Comprehension}
As a challenging task in NLP, text comprehension is one of the key problems in artificial intelligence, which aims to read and comprehend a given text, and then answer questions or make inference based on it. These tasks require a comprehensive understanding of natural languages and the ability to do further inference and reasoning. We focus on two types of text comprehension, document-based question-answering (Table \ref{tab:mrc_example}) and textual entailment (Table \ref{tab:rte_example}). Textual entailment aims for a deep understanding of text and reasoning, which shares the similar genre of machine reading comprehension, though the task formations are slightly different. 

In the last decade, the MRC tasks have evolved from the early cloze-style test \citep{hill2015goldilocks,hermann2015teaching,zhang2018char,zhang2018SubMRC} to span-based answer extraction from passage \citep{Rajpurkar2016SQuAD,Rajpurkar2018Know}. The former has restrictions that each answer should be a single word  in the document and the original sentence without the answer part is taken as the query. For the span-based one, the query is formed as questions in natural language whose answers are spans of texts. Various attentive models have been employed for text representation and relation discovery, including Attention Sum Reader \citep{kadlec2016text}, Gated attention Reader \citep{Dhingra2017Gated} and Self-matching Network \citep{Wang2017Gated}.

With the release of the large-scale span-based datasets \citep{Rajpurkar2016SQuAD,Joshi2017TriviaQA,Rajpurkar2018Know}, which constrain answers to all possible text spans within the reference document, researchers are investigating the models with more logical reasoning and content understanding \citep{wang2018multi}. Recently, language models also show their remarkable performance in reading comprehension \citep{devlin2018bert,Peters2018ELMO}.

For the other type of text comprehension, natural language inference (NLI) is proposed to serve as a benchmark for natural language understanding and inference, which is also known as recognizing textual entailment (RTE). In this task, a model is presented with a pair of sentences and asked to judge the relationship between their meanings, including entailment, neutral and contradiction. \citet{Bowman2015A} released Stanford Natural language Inference (SNLI) dataset, which is a high-quality and large-scale benchmark, thus inspiring various significant work.

\begin{table}
	\centering \scriptsize
	
	{
		\begin{tabular}{l|p{6cm}}
			\hline
			\hline
			{\bf Passage }
			& There are three major types of rock: igneous, sedimentary, and metamorphic. The rock cycle is an important concept in geology which illustrates the relationships between these three types of rock, and magma. When a rock crystallizes from melt (magma and/or lava), it is an igneous rock. This rock can be weathered and eroded, and then redeposited and lithified into a sedimentary rock, or be turned into a metamorphic rock due to \textcolor{red}{heat and pressure} that change the mineral content of the rock which gives it a characteristic fabric. The sedimentary rock can then be subsequently turned into a metamorphic rock due to heat and pressure and is then weathered, eroded, deposited, and lithified, ultimately becoming a sedimentary rock. Sedimentary rock may also be re-eroded and redeposited, and metamorphic rock may also undergo additional metamorphism. All three types of rocks may be re-melted; when this happens, a new magma is formed, from which an igneous rock may once again crystallize. \\ 
			\hline
			{\bf Question} & What changes the mineral content of a rock? \\
			\hline
			{\bf Answer} & heat and pressure. \\
			\hline
			\hline
		\end{tabular}
	}
	\caption{\label{tab:mrc_example} A machine reading comprehension example.}
\end{table}

\begin{table}
	\centering \scriptsize
	
	{
		\begin{tabular}{l|l|l}
			\hline
			\hline
			Premise & A man parasails in the choppy water. & Label\\
			\hline
			\multirow{3}{*}{Hypo.} & The man is competing in a competition. & Neutral \\
			& The man parasailed in the calm water. &  Contra. \\
			&  The water was choppy as the man parasailed. & Entailment \\
			\hline
			\hline
		\end{tabular}
	}
	\caption{\label{tab:rte_example} A textual entailment example.}
\end{table}

Most of existing NLI models apply attention  mechanism  to  jointly  interpret  and  align the premise and hypothesis,  while  transfer  learning  from
external knowledge is popular recently. Notably, \citet{Chen2017Enhanced}  proposed  an  enhanced  sequential  inference model (ESIM), which employed recursive architectures in both local inference modeling and
inference composition, as well as syntactic
parsing information, for a sequential inference model. ESIM is simple with satisfactory performance, and thus is widely chosen as the baseline model. \citet{Mccann2017Learned} proposed  to
transfer  the  LSTM  encoder  from  the  neural
machine translation (NMT) to the NLI task to contextualize word vectors. \citet{pan2018discourse} transferred the  knowledge learned from the discourse marker prediction task
to the NLI task to augment the semantic representation.

\subsection{Semantic Role Labeling}
Given a sentence, the task of semantic role labeling is dedicated to recognizing the semantic relations between the predicates
and the arguments. For example, given the sentence, \emph{Charlie sold a book to Sherry last week}, where the target verb (predicate) is \emph{sold}, SRL system yields the following outputs,
\begin{spacing}{1.3}
\end{spacing}
[$_{ARG0}$ Charlie] [$_{V}$ sold] [$_{ARG1}$ a book] 

[$_{ARG2}$ to Sherry] [$_{AM-TMP}$ last week].
\begin{spacing}{1.3}
\end{spacing}
\noindent where $ARG0$ represents the  seller (agent), $ARG1$ represents the thing sold (theme), $ARG2$ represents the buyer (recipient), $AM-TMP$ is an adjunct indicating the timing of the action and $V$ represents the predicate.

Recently, SRL has aroused much attention from researchers and has been applied in many NLP tasks \citep{Mihaylov2016Discourse,Shi2016Knowledge,Yih2016The}. SRL task is generally formulated as multi-step classification subtasks in pipeline systems, consisting of predicate identification, predicate disambiguation, argument identification and argument classification. Most previous SRL approaches adopt a pipeline framework to handle these subtasks one after another. Notably, \citet{Gildea2002Automatic}  devised the first automatic semantic role labeling system based on FrameNet. Traditional systems relied on sophisticated handcraft features or some declarative constraints, which suffer from poor efficiency and generalization ability. A recently tendency for SRL is adopting neural networks methods thanks to their significant success in a wide range of applications. The pioneering work on building an end-to-end neural system was presented by \citep{zhou-xu2015}, applying an 8 layered LSTM model, which takes only original text information as input feature without using any syntactic knowledge, outperforming the previous state-of-the-art system. \citet{He2017Deep} presented a deep highway BiLSTM architecture with constrained decoding, which is simple and effective, enabling us to select it as our basic semantic role labeler. These studies tackle argument identification and argument classification in one shot. Inspired by recent advances, we can easily integrate semantics into text comprehension.

\iffalse
\citet{He2018Jointly} proposed an end-to-end approach for jointly predicting all predicates, arguments spans, and the relations between them. \citet{Marcheggiani2017A} also proposed a simple and accurate neural model using predicate specific  encodings  of a sentence and use them to predict arguments of the corresponding predicate. Without using any syntactic information, their approach achieved the state-of-the-art result on the CoNLL-2009 dataset. \citet{He2018Syntax} presented a simple and effective neural model  for dependency-based  SRL,  incorporating syntactic information with the proposed extended k-order pruning algorithm.  
\fi

\section{Semantic Role Labeling for Text Comprehension}

\begin{figure}
	\centering
	\includegraphics[width=0.36\textwidth]{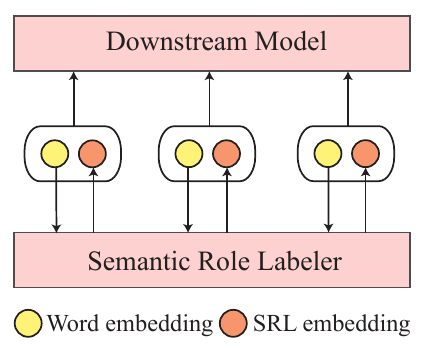}
	\caption{\label{fig:framework}Overview of the semantic learning framework.}
\end{figure}

For both downstream text comprehension tasks, we consider an end-to-end model as well as the semantic learning model. The former may be regarded as downstream model of the latter. Thus, our semantics augmented model will be an integration of two end-to-end models through simple embedding concatenation as shown in Figure \ref{fig:framework}. 

In detail, we apply semantic role labeler to annotate the semantic tags (i.e. predicate, argument) for each token in the input sequence so that explicit contextual semantics can be directly introduced, and then the input sequence along with the corresponding semantic role labels is fed to downstream models. We regard the semantic signals as SRL embeddings and employ a lookup table to map each label to vectors, similar to the implementation of word embedding. For each word $x$, a joint embedding $e^{j}(w)$ is obtained by the concatenation of word embedding $e^{w}(x)$ and SRL embedding $e^{s}(x)$,
\begin{align*}	
e^{j}(w) = e^{w}(x) \oplus e^{s}(x)
\end{align*}
where $\oplus$ is the concatenation operator.
The downstream model is task-specific. In this work, we focus on the textual entailment and machine reading comprehension, which will be discussed latter.

\begin{figure*}
	\centering
	\includegraphics[width=0.88\textwidth]{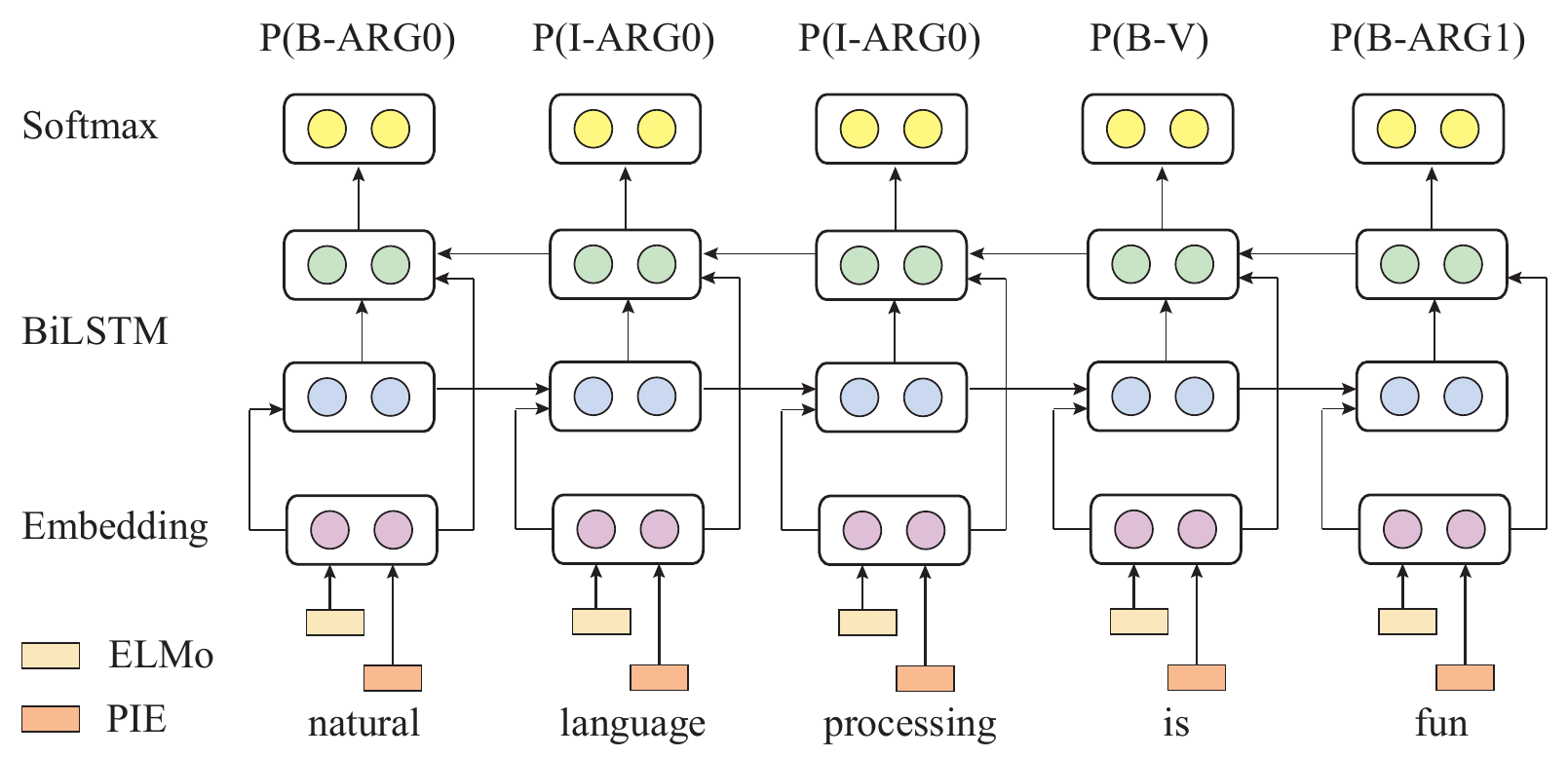}
	\caption{\label{fig:srl_module}Semantic role labeler.}
\end{figure*}

\subsection{Semantic Role Labeler}
Our concerned SRL task includes two subtasks: predicate identification and argument labeling. While the CoNLL-2005 shared task assumes gold predicates as input, this information is not available in many applications, which requires us to identify the predicates for a input sentence at the very beginning. Thus, our SRL module has to be end-to-end, predicting all predicates and corresponding arguments in one shot.

For predicate identification, we use spaCy\footnote{https://spacy.io/} to tokenize the input sentence with part-of-speech (POS) tags and the verbs are marked as the binary predicate indicator for whether the word is the verb for the sentence.

Following \citep{He2017Deep}, we model SRL as a span tagging problem\footnote{Actually, the easiest way to deal with segmentation or sequence labeling problems is to transform them into raw labeling problems. A standard way to do this is the \emph{BIO} encoding, representing a token at the beginning, interior, or outside of any span, respectively.} and use an 8-layer deep BiLSTM with forward and backward directions interleaved. Different from the baseline model, we replace the GloVe embeddings with ELMo representations\footnote{The ELMo representation is obtained from \url{https://allennlp.org/elmo}. We use the original one for this work whose output size is 512.} due to the recent success of ELMo in NLP tasks \citep{Peters2018ELMO}. 

In brief, the implementation of our SRL is a series of stacked interleaved LSTMs with highway connections. The inputs are embedded sequences of words concatenated with a binary indicator containing whether a word is the verbal predicate. Additionally, during inference, Viterbi decoding is applied to accommodate valid BIO sequences. The details are as follows.

\paragraph{Word Representation}
The word representation of our SRL model is the concatenation of two vectors: an ELMo embedding $\bm{e}^{(l)}$ and predicate indicator embedding (PIE) $\bm{e}^{(p)}$. ELMo is trained from the internal states of a deep bidirectional language model (BiLM), which is pre-trained on a large text corpus with approximately 30
million sentences \citep{Chelba2014One}. Besides, following \citep{li2019dependency} who shows the predicate-specific feature is helpful in promoting the role labeling, we employ a predicate indicator embedding $\bm{e}^{(p)}$ to mark whether a word is a predicate when predicting and labeling the arguments. The final word representation is given by $\bm{e} = \bm{e}^{(l)} \oplus \bm{e}^{(p)}$, where $\oplus$ is the concatenation operator. The downstream model will take such a joint embedding as input for specific task.

\paragraph{Encoder} 
As commonly used to model the sequential input, BiLSTM is adopted for our sentence encoder. By incorporating a stack of distinct LSTMs, BiLSTM processes an input sequence in both forward and backward directions. In this way, the BiLSTM encoder provides the ability to incorporate the contextual information for each word. 

Given a sequence of word representation $S=\{\bm{e}_1, \bm{e}_2, \cdots, \bm{e}_n\}$ as input, the hidden state $\bm{h}=\{\bm{h}_1, \bm{h}_2, \cdots, \bm{h}_n\}$ is encoded by BiLSTMs layer where each LSTM uses highway connections between layers and variational recurrent dropout. The encoded representation is then projected using a final dense layer followed by a softmax activation to form a distribution over all possible tags. The predicted semantic role Labels are defined in PropBank \citep{Palmer2005The} augmented with B-I-O tag set to represent argument spans. 

\paragraph{Model Implementation}
The training objective is to maximize the logarithm of the likelihood of the tag sequence, and we expect the correct output sequence matches with,
\begin{equation}
y^{*} = \mathop{argmax}\limits_{\widetilde{y}\in C}s(x,\widetilde{y})
\end{equation}
where $\mathbf{C}$ is candidate label set.

Our semantic role labeler is trained on English \emph{OntoNotes v5.0} dataset \citep{pradhan2013towards}  for the CoNLL-2012 shared task, achieving an F1 of 84.6\%\footnote{This result is comparable with the state-of-the-art \citep{li2019dependency}.} on the test set. At test time, we perform Viterbi decoding to enforce valid spans using BIO constraints\footnote{The BIO format requires argument spans to begin with a B tag.}. For the following evaluation, the default dimension of SRL embeddings is 5 and the case study concerning the dimension is shown in the subsection \emph{dimension of SRL Embedding}.

The model is run forward for every verb in the sentence.  In some cases there is more than one predicate in a sentence, resulting in various semantic role sets whose number is equal to the number of predicates. For convenient downstream model input, we need to ensure the word and the corresponding label are matched one-by-one, that is, only one set for a sentence. To this end, we select the corresponding BIO sets with the most non-O labels as the semantic role labels. For sentences with no predicate, we directly assign \emph{O} labels to each word in those sentences. 

\subsection{Text Comprehension Model}
\paragraph{Textual Entailment} Our basic TE model is the reproduced Enhanced Sequential Inference Model (ESIM) \citep{Chen2017Enhanced} which is a widely used baseline model for textual entailment. ESIM employs a BiLSTM to encode the premise and hypothesis, followed by an attention layer, a local inference layer, an inference composition layer. Slightly different from \citep{Chen2017Enhanced}, we do not include extra syntactic parsing features and directly replace the pre-trained Glove word embedding with ELMo which are completely character based. Our SRL embedding is concatenated with ELMo embeddings and the joint embeddings are then fed to the BiLSTM encoders. 

\paragraph{Machine Reading Comprehension}

Our baseline MRC model is an enhanced version of Bidirectional Attention Flow \citep{Seo2016Bidirectional} following \citep{clark2018simple}. The token embedding is the concatenation of pre-trained GloVe word vectors, a character-level embedding from a convolutional neural network with max-pooling and pre-trained ELMo embeddings \citep{Peters2018ELMO}. Our semantics enhanced model takes input of concatenating the token embedding with SRL embeddings. The embeddings of document and question are passed through a shared bi-directional GRU, followed by a BiDAF attention \citep{Seo2016Bidirectional}. The contextual document and question representations are then passed to a residual self-attention layer. The above model is denoted as \emph{ELMo}. 
Table \ref{tab:squad} shows the results on SQuAD MRC task\footnote{For BERT evaluation, we only use SQuAD training set instead of joint training with other datasets to keep the model simplicity. Since the test set of SQuAD is not publicly available, our evaluations are based on dev set.}. The SRL embeddings give substantial performance gains over all the strong baselines, showing it is also quite effective for more complex document and question encoding.

\begin{table}[H]
	\centering
	{
		\begin{tabular}{l c}
			\hline
			
			\hline
			Model  & Accuracy (\%) \\ 
			\hline
			Deep Gated Attn. BiLSTM 	 & 85.5 \\
			Gumbel TreeLSTM  	 & 86.0 \\
			Residual stacked  	 & 86.0 \\
			Distance-based SAN  	 	& 86.3 \\
			\hline
			BCN + CoVe + Char   	 & 88.1 \\
			DIIN 		 & 88.0 \\
			DR-BiLSTM   	 & 88.5 \\
			CAFE  	 	 & 88.5\\
			MAN  	 	 & 88.3\\
			KIM  	 	 & 88.6 \\
			DMAN    & 88.8 \\
			ESIM + TreeLSTM   	   & 88.6\\
			ESIM + ELMo   	 	 & 88.7 \\
			DCRCN    & 88.9 \\
			LM-Transformer   & 89.9 \\
			MT-DNN$\dagger$ & 91.1\\
			\hline
			Baseline (ELMo)     &88.4 \\
			\textbf{+ SRL}   &  89.1 \\
			Baseline (BERT$_\text{BASE}$)    &89.2\\
			\textbf{+ SRL}   &  89.6 \\
			Baseline (BERT$_\text{LARGE}$)      & 90.4 \\
			\textbf{+ SRL}    & \textbf{91.3} \\
			\hline
			
			\hline
		\end{tabular}
	}
	
	\caption{\label{tab:snli} Accuracy on SNLI test set. Models in the first block are sentence encoding-based. The second block embodies the joint methods while the last block shows our SRL based model. All the results except ours are from the SNLI Leaderboard. Previous state-of-the-art model is marked by $\dagger$. Since ensemble systems are commonly integrated with multiple heterogeneous models and resources, we only show the results of single models to save space though our single model also outperforms the ensemble models.}
\end{table}

\section{Evaluation}
In this section, we evaluate the performance of SRL embeddings on two kinds of text comprehension tasks, \emph{textual entailment} and \emph{reading comprehension}. Both of the concerned tasks are quite challenging, and could be even more difficult considering that the latest performance improvement has been already very marginal. However, we present the semantics enhanced solution instead of heuristically stacking network design techniques to give further advances. In our experiments, we basically follow the same hyper-parameters for each model as the original settings from their corresponding literatures \citep{Peters2018ELMO,Chen2017Enhanced,clark2018simple} except those specified (e.g. SRL embedding dimension). For both of the tasks, we also report the results by using pre-trained BERT \citep{devlin2018bert} as word representation in our baseline models \footnote{We use the last layer of BERT output. Since BERT is in subword-level while semantics role labels are in word-level, to use BERT in conjunction with our SRL embeddings, we need to keep them aligned. Therefore, we use the BERT embedding for the first subword of each word, which is slightly different from the original \href{https://github.com/google-research/bert}{BERT}. }. The hyperparameters were selected using the Dev set, and the reported Dev and Test scores are averaged over 5 random seeds using those hyper-parameters.

\subsection{Textual Entailment} 

Textual entailment is the task of determining whether a \emph{hypothesis} is \emph{entailment, contradiction} and \emph{neutral}, given a \emph{premise}. The Stanford Natural Language Inference (SNLI) corpus \citep{Bowman2015A} provides approximately 570$k$ hypothesis/premise pairs. We evaluate the model performance in terms of accuracy.

Results in Table \ref{tab:snli} show that SRL embedding can boost the ESIM+ELMo model by +0.7\% improvement. With the semantic cues, the simple sequential encoding model yields substantial gains, and our single BERT$_\text{LARGE}$ model also achieves a new state-of-the-art, even outperforms all the ensemble models in the leaderboard\footnote{Since March 24th, 2019. The leaderboard is here: https://nlp.stanford.edu/projects/snli/. }. This would be owing to more accurate and fine-grained information from effective explicit semantic cues.

\begin{table}
	\centering
	{
		\begin{tabular}{l c c}
			\hline
			
			\hline
			Model& Dev & Test\\
			\hline
			\textbf{Our model} &\textbf{89.11} & \textbf{89.09} \\
			\hline
			-ELMo & 88.51   & 88.42 \\
			-SRL &  88.89   &  88.65 \\	
			-ELMo -SRL &  88.39  & 87.96 \\
			\hline
			
			\hline
		\end{tabular}
	}
	
	\caption{\label{tab:ablation} Ablation study. Since we use ELMo as the basic word embeddings, we replace ELMO with 300D GloVe embeddings for the case \emph{-ELMo}.}
\end{table}
To evaluate the contributions of key factors in our
method, a series of ablation studies are performed
on the SNLI dev and test set. The results are in Table \ref{tab:ablation}. We observe both SRL and ELMo embeddings contribute to the overall performance. Note that ELMo is obtained by deep bidirectional language with 4,096 hidden units on a large-scale corpus, which requires long training time with 93.6 million parameters. The output dimension of ELMo is 512. Compared with the massive computation and high dimension, SRL embedding is much more convenient for training and much easier for model integration, giving the same level of performance gains.

\subsection{Machine Reading Comprehension} 

To investigate the effectiveness of the SRL embedding in conjunction with more complex models, we conduct experiments on machine reading comprehension tasks. The reading comprehension task can be described as a triple $<D, Q, A>$, where $D$ is a document (context), $Q$ is a query over the contents of $D$, in which a span is the right answer $A$.  

As a widely used benchmark dataset for machine reading comprehension, the Stanford Question Answering Dataset (SQuAD) \citep{Rajpurkar2016SQuAD} contains 100$k$+ crowd sourced question-answer pairs where the answer is a span in a given Wikipedia paragraph. Two official metrics are selected to evaluate the model performance: Exact Match (EM) and a softer metric F1 score, which measures the weighted average of the precision and recall rate at a character level. Our baseline includes MQAN \citep{Mccann2018The} for single task and multi-task with SRL, BiDAF+ELMo \citep{Peters2018ELMO}, R.M. Reader and BERT \citep{devlin2018bert}.

\begin{table}
	\centering
	
	{
		\begin{tabular}{l c c c}
			\hline
			
			\hline
			\textbf{Model}& \textbf{EM}& \textbf{F1} & \textbf{RERR} \\
			\hline
			\multicolumn{4}{c}{\emph{Published}} \\
			MQAN$_\text{single-task}$ & - & 75.5 &- \\
			MQAN$_\text{multi-task}$ & - & 74.3 &- \\
			BiDAF+ELMo & - &85.6 & -\\
			R.M. Reader & 78.9 & 86.3 & - \\
			BERT$_\text{BASE}$ & 80.8 & 88.5 & - \\
			BERT$_\text{LARGE}$$\dagger$ & 84.1 & 90.9 & - \\
			\hline
			\multicolumn{4}{c}{\emph{Our implementation}} \\
			Baseline (ELMo)& 77.5 & 85.2 & -\\
			\textbf{+SRL} &  \textbf{78.5} & \textbf{86.0} & \textbf{5.4\%}\\
			\hline
			Baseline (BERT$_\text{BASE}$)  &  81.3 &  88.5 & -\\
			\textbf{+SRL} &  \textbf{81.7} & \textbf{88.8} & \textbf{ 2.6\%}\\
			\hline
			Baseline (BERT$_\text{LARGE})$  & 84.2  & 90.9 & -\\
			\textbf{+SRL} &  \textbf{84.5} & \textbf{91.2} & \textbf{3.3\%}\\
			\hline
		\end{tabular}
	}
	
	\caption{\label{tab:squad} Exact Match (EM) and F1 scores on SQuAD dev set. RERR is short for relative error rate reduction of our model to the baseline evaluated on F1 score. Previous state-of-the-art model is marked by $\dagger$.}
\end{table}

Table \ref{tab:squad} shows the results\footnote{Since the test set of SQuAD is not publicly available, our evaluations are based on dev set.}. The SRL embeddings give substantial performance gains over all the strong baselines, showing it is also quite effective for more complex document and question encoding.

\section{Case Studies}
From the above experiments, we see our semantic learning framework works effectively and the semantic role labeler boosts model performance, verifying our hypothesis that semantic roles are critical for text understanding. Though the semantic role labeler is trained on a standard benchmark dataset, \emph{Ontonotes}, whose source ranges from news, conversational telephone speech, weblogs, etc., it turns out to be generally useful for text comprehension from probably quite different domains in both textual entailment and machine reading comprehension. To further evaluate the proposed method, we conduct several case studies as follows.

\begin{figure}
	\centering
	\includegraphics[width=0.42\textwidth]{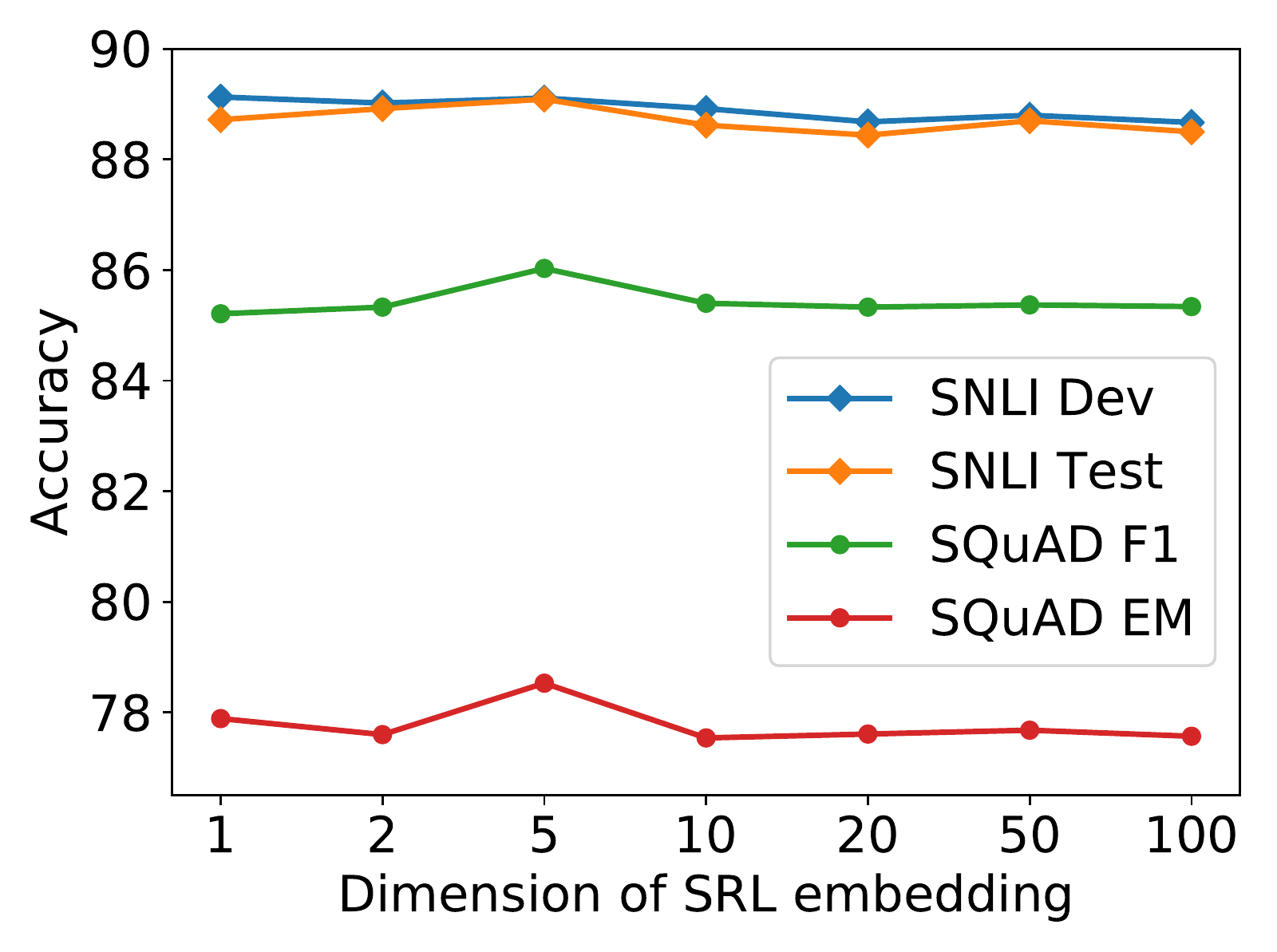}
	\caption{\label{fig:squad_dim} Results on SNLI and SQuAD with different SRL embedding dimensions. }
\end{figure}

\begin{table}
	\centering
	{
		\begin{tabular}{l c c}
			\hline
			
			\hline
			Model& Dev & Test\\
			\hline
			Baseline &   88.89   &  88.65  \\
			\hline
			\textbf{Word + SRL} &\textbf{89.11} & \textbf{89.09} \\
			Word + POS & 88.90 &  88.68 \\	
			Word + NE & 89.14 &  88.51 \\
			\hline
			
			\hline
		\end{tabular}
	}
	
	\caption{\label{tab:snli_comp} Comparison with different NLP tags. }
\end{table}

\subsection{Dimension of SRL Embedding}
The dimension of embedding is a critical hyper-parameter in deep learning models that may influence the performance. Too high dimension would cause severe over-fitting issues while too low dimension would also cause under-fitting results. To investigate the influence of the dimension of SRL embeddings, we change the dimension in the intervals [1, 2, 5, 10, 20, 50, 100]. Figure \ref{fig:squad_dim} shows the results. We see that 5-dimension SRL embedding gives the best performance on both SNLI and SQuAD datasets. 

\subsection{Comparison with POS/NER Tags}

The study of computational linguistics is a critical part in NLP \citep{zhou2019head,li-etal-2018-joint-learning}. In particular, Part-of-speech (POS) and named entity (NE) tags have been broadly used in various tasks. To make comparisons, we conduct experiments on SNLI with modifications on label embeddings using tags of SRL, POS and NE, respectively. Results in Table \ref{tab:snli_comp} show that SRL gives the best result, showing semantic roles contribute to the performance, which also indicates that semantic information matches the purpose of NLI task best.

\section{Conclusion}

This paper presents a novel semantic learning framework for fine-grained text comprehension and inference. We show that our proposed method is simple yet powerful, which achieves a significant improvement over strong baseline models, including those which have been enhanced by the latest BERT. This work discloses the effectiveness of explicit semantics in text comprehension and inference and proposes an easy and feasible scheme to integrate explicit contextual semantics in neural models.  A series of detailed case studies are employed to analyze the adopted robustness of the semantic role labeler. Different from most recent works focusing on heuristically stacking complex mechanisms for performance improvement, this work is to shed some lights on fusing accurate semantic signals for deeper comprehension and inference. 
\bibliography{paclic33}

\end{document}